\documentclass[a4paper,10pt]{article}
\usepackage{amsmath,amsfonts,amssymb}
\usepackage{algorithmic}
\usepackage{subfig}
\usepackage{url}
\usepackage{graphicx}

\usepackage[left=1.75cm,right=1.75cm]{geometry}

\usepackage[numbers]{natbib}

\usepackage{bm}
\newcommand{\emg}{{\Phi}}
\newcommand{\p}{\bm{p}}
\newcommand{\x}{\bm{x}}

\usepackage{xcolor}

\usepackage{multirow}

\usepackage{siunitx}

\usepackage{array}
\newcolumntype{R}[1]{>{\raggedleft\arraybackslash}m{#1}}

\usepackage{authblk}

\begin{document}
	\title{Estimation of Motor Unit Parameters from Surface Electromyograms using an Informed Autoencoder}

	\author[1]{Kaja Balzereit}
	\author[1]{Malte Mechtenberg}
	\author[1]{Axel Schneider}
	
	\affil[1]{Hochschule Bielefeld, University of Applied Sciences and Arts, Institute for System Dynamics and Mechatronics}
	
	\date{}
	
	\maketitle
	
	\begin{abstract}
		Motor unit parameters such as the innervation zone centre or the conduction velocity of the electrical potential harbour the potential to improve the fidelity of neuromechanical models used for movement and force prediction.
		Determining these parameters in a non-invasive way is challenging, as they are subject-specific and may vary with muscle contraction.
		Existing work on the estimation of motor unit parameters mainly relies on white-box modelling and therefore requires substantial manual modelling effort.
		This work targets the simultaneous estimation of multiple subject-specific motor unit parameters from electromyography (EMG) recordings measured non-invasively at the skin surface.
		This results in an inverse problem with a nonlinear loss function.
		To address this problem, an informed autoencoder is developed.
		This autoencoder reconstructs the surface EMG recordings while learning the parameters in its latent space and adhering to physical laws that relate the parameters to the EMG signals.
		In experiments on synthetic data, innervation zone centres are estimated with a mean absolute error of $2.5989 \,\si{\milli\meter}$, and conduction velocities of the electric potential are estimated with a mean absolute error of $0.1697 \,\si{\meter\per\second}$.
		These results demonstrate the plausibility of this novel approach, which enables the simultaneous estimation of several motor unit parameters while reducing manual modelling effort through the integration of data-driven machine learning.
	\end{abstract}	
	
	\section{Introduction}
	
	For biosignal-based prediction of limb movements, a neuromechanical (white-box) model of the relevant joints -- including submodels of muscles, tendons, and their innervation -- is essential \cite{Grimmelsmann2023,Mechtenberg2022}.
	If parts of the submodels are unknown or parameters are difficult to estimate, one resorts to purely data-driven (black-box) approaches that disregard physiological knowledge.
	
	Biosignals can be acquired non-invasively either centrally, via electroencephalography (EEG, e.g. in brain-computer interfaces), or peripherally, via surface electromyography (sEMG) recorded with electrodes on the skin.
	For instance, prediction of elbow joint motion can be based on sEMG activity from the principal flexors (e.g. biceps brachii, brachialis, brachioradialis) and extensors (e.g. triceps brachii, anconeus) \cite{Grimmelsmann2023}, and hand gestures can be recognised from sEMG and acceleration recordings \cite{Pan.2022}.
	A key physiological feature is the electromechanical delay -- the measurable lag between the propagation of muscle fibre action potentials and the resulting force production or joint movement \cite{Cavanagh1979} -- typically on the order of several tens up to approximately 100 ms \cite{Vos1991}.
	This delay renders sEMG intrinsically predictive of imminent force generation \cite{Ce2013}.
	However, it is influenced by various physiological parameters \cite{Ce2013,Moreno2025_Electromech_Delay_Stimulation}
	
	In neuromechanical modelling, each muscle-tendon unit (MTU) comprises a muscle (or muscle head) and its associated tendons spanning a joint \cite{Zajac1989}.
	Most movement prediction models rely solely on total MTU length.
	However, during contraction, both muscle and tendon lengths vary under load, and resolving their individual contributions is critical for accurate force prediction \cite{Zajac1989}, especially under dynamic force conditions \cite{Grimmelsmann2023,Buchanan2004}.
	High-density sEMG sensor arrays enable observation of the spatio-temporal propagation of motor unit action potentials along muscle fibres.
	These action potentials originate at the innervation zone -- where the neuromuscular junction of a motor neuron interfaces with a muscle fibre -- and propagate bidirectionally along the fibre.
	By tracking their propagation across multiple electrodes, conduction velocity can be estimated and the origin of excitation inferred retrospectively \cite{mechtenberg2023method}.
	
	This principle extends to the aggregate (interference) EMG signal, allowing identification of the effective innervation centre of the active motor unit population.
	Importantly, the apparent position of this centre shifts mechanically during muscle contraction.
	Tracking this displacement using sEMG alone provides a means to estimate internal muscle deformation \cite{Mechtenberg2024}.
	Consequently, monitoring the innervation centre offers a practical route to partition total MTU length into its muscle and tendon components, thereby improving the fidelity of neuromechanical models used for movement and force prediction.

	\subsection{Problem Description}
	
	The problem addressed in this article is the simultaneous estimation of parameters specifying motor units such as the innervation centre and the conduction velocity of the electrical potential from sEMG measurements.
	If the parameters of individual motor units can be estimated from sEMG, it would be possible to use these motor unit parameters for further muscle movement analysis or motor unit identification and decomposition.
	One challenge that arises when estimating those parameters is their high number:
	Each muscle consists of several motor units, which in turn consist of multiple muscle fibres.
	For example, for young men the biceps brachii consists of approximately $250 000$ muscle fibres \cite{Klein.2003}.
	The activation of each of these muscle fibres results in an electrical field, which strength depends on parameters specific to each muscle fibre.
	EMG measurements recorded at skin level are the result of the superposition of the electrical fields resulting from all muscle fibres.
	Consequently, the EMG measurements are defined by $250\, 000 \cdot q$ parameters, where $q \in \mathbb{N}$ is the number of parameters defining a single muscle fibre.
	
	There are several approaches for estimating the innervation zone centre from sEMG recordings.
	\citet{Beck2012-fj} compared three different approaches to estimate the electrode closest to the innervation zone:
	(i) cross-correlation on adjacent EMG channels, (ii) minimum amplitude, and (iii) maximum centre frequency.
	Their evaluation showed that the cross-correlation-based estimation was the most accurate among these three.
	Nevertheless, their approach focuses on estimating the closest electrode and thus, the accuracy of this approach is limited by the inter electrode distance (IED).
	
	\citet{Mechtenberg2024} presented an approach on the estimation of the innervation zone centre from sEMG based on tracking the bidirectional propagation of motor unit potentials, determining candidates for the innervation point by calculating the intersection of lines tracing the motor unit potential back in time, and combining this with a clustering algorithm that determines the centre of these intersection points.
	Similar work has been published by \citet{Marateb2016-ii}, who generated images from sEMG and analysed the spatio-temporal propagation of motor unit potentials.
	This propagation analyses determines the innervation zone centre and the conduction velocity of the electrical potential.
	
	All of these approaches are white-box approaches specialised in identifying the location of innervation zone or the innervation zone centre, respectively, and in the case of Marateb \cite{Marateb2016-ii}, also the conduction velocity.
	Nevertheless, extending these approaches to predict further parameters such as the length of the muscle fibres or the position of the muscle relative to the electrode array would require domain knowledge and further manual modelling efforts.
	
	This article presents a versatile learning approach that allows for the prediction of several motor unit parameters whilst adhering to physical principles.
	As the interpretability of the estimated parameters and their adherence to physical laws is crucial for physiological applications, a grey-box approach, specifically an informed machine learning approach \cite{Rueden.2021,Karniadakis.2021}, is employed instead of a black-box approach.
	This approach combines physiological knowledge with data-driven learning, ensuring the consideration of physical laws at all times and preserving the interpretability of the estimated parameters.
	
	Another challenge that arises is the lack of ground-truth values, as the acquisition of true values of motor unit parameters is difficult.
	Consequently, the application of supervised learning is infeasible.
	sEMG recordings, in contrast, implicitly contain information about the motor unit parameters, as the electrical potential recorded at skin level stems from the superposition of the electrical fields resulting from several muscle fibres, which are characterized by the sought parameters.
	Therefore, an informed autoencoder architecture is created that reconstructs sEMG recordings whilst learning the parameters to be estimated in its latent space.

	\subsection{Contribution and Outline of the Article}
	
	The contribution of this work is as follows:
	(i) A novel approach to the estimation of motor unit parameters, using EMG recordings that are non-invasively measured at skin level, is introduced.
	Given measurements from one individual, the approach estimates parameters specific to this individual.
	For this purpose, an informed neural network architecture, inspired by the Encoder-X architecture \cite{wang2021encoder}, is employed, which allows combining a forward model \cite{mechtenberg2023method}, created from domain knowledge, with a data-driven approach.
	(ii) The experimental evaluation demonstrates the feasibility of the approach.
	The innervation zone centre, i.e. the mean value of the innervation points of all muscle fibres of a motor unit, is estimated with a mean absolute error of $2.5989 \,\si{\milli\meter}$, and the conduction velocities, i.e. the velocities at which the electrical potential propagates through the tissue, are estimated with a mean absolute error of $0.1697 \,\si{\meter\per\second}$.
	
	This article is structured as follows:
	In Section \ref{sec:method}, the approach based on an informed autoencoder reconstructing sEMG measurements is presented.
	This approach is applied to various scenarios for the estimation of the innervation zone and the conduction velocity, and the results are given in Section \ref{sec:experiments}.
	Finally, the work is concluded and an outlook on future work is provided in Section \ref{sec:conclusion}.

	\section{Solution Approach} \label{sec:method}
	
	As the ground truth values of motor unit parameters are difficult to acquire, many approaches rely on sEMG recordings to estimate motor unit parameters as these recordings implicitly contain information about the parameters.
	The computation of motor unit parameters from sEMG recordings, however, remains challenging, as no direct function allowing for the determination of parameters from sEMG recordings is known.
	However, (non-linear) forward models \cite{Klotz.2020,Duchene.2000,mechtenberg2023method} allowing for the calculation of the electric potential field, given motor unit parameters, are known.
	Using a forward model, the estimation of parameters from sEMG recordings leads to a non-linear inverse problem.
	Given a set of sEMG recordings $\Psi$ and the forward model, the goal is to find those parameters $\hat{\p}^\mathrm{min}$ that best explain the sEMG recordings, so $\hat{\p}^\mathrm{min}$ shall minimise the discrepancy (given a loss function) between the prediction from the model and the recordings.
	The solution of this inverse problem requires the minimisation of a residual function, i.e. the discrepancy between a prediction of the sEMG (using a forward model created from domain knowledge) and measured sEMG recordings.
	However, this residual function exhibits a multitude of local minima.
	
	Neural networks have been shown to allow for the automatic extraction of features from data while omitting the need for an individual parametrization \cite{lecun2015deep}.
	For unsupervised learning tasks, autoencoders \cite{hinton2006reducing} are a well-known architecture for dimensionality reduction and for estimating parameters, for example of polynomials \cite{wang2021encoder} or ODEs \cite{kisbenedek2024autoencoder}.
	An autoencoder, originally presented by \citet{hinton2006reducing}, consists of two parts: an encoder and a decoder \cite{Baldi.2012}.
	The encoder takes data as input and maps it to a latent space, which often has a lower dimension than the input data.
	These latent variables form the input for the decoder, which aims to reconstruct the input data.
	Originally, both the encoder and the decoder are built using multi-layer neural networks.
	For training the weights of these networks, the reconstruction error, i.e. the difference between the input data and the output of the decoder, is used \cite{hinton2006reducing}.
	
	\begin{figure*}
		\includegraphics[width=\linewidth]{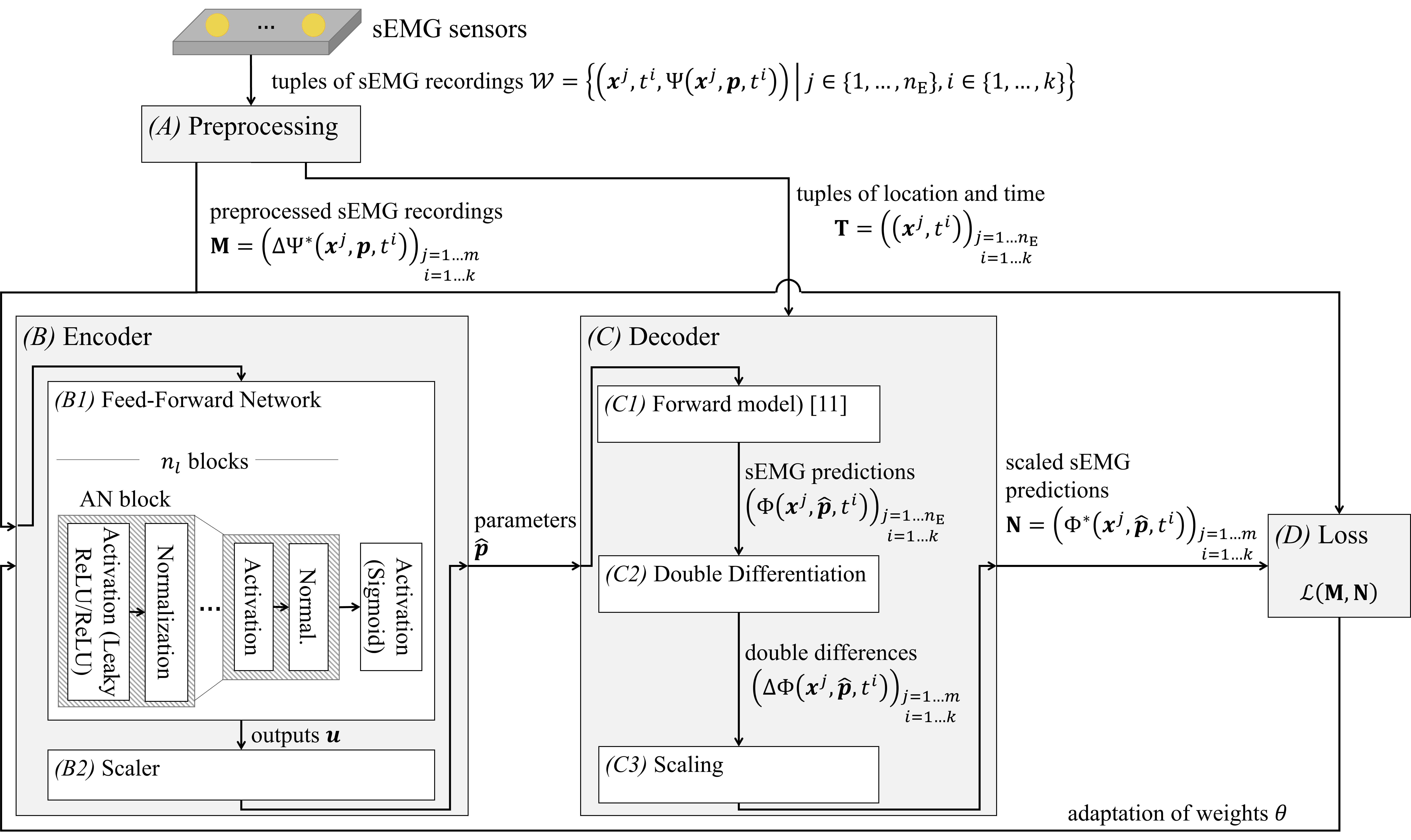}
		\caption{
			Basic idea of the proposed approach: 
			The input data is formed by a set $\mathcal{W}$ of tuples of sEMG recordings $\Psi(\p)$ of a single motor unit, resulting from the superposition of $n \in \mathbb{N}$ ($n$ might be unknown) electrical fields from $n$ muscle fibres, at several times $t^i$ and locations $\x^j$ of the electrodes, measured by sEMG sensors.
			These recordings first are preprocessed \textit{(A)}.
			The preprocessed sEMG recordings, denoted by $\bm{\mathrm{{M}}} = \left(\Delta {\Psi}^{*}(\x^j, \p, t^i)\right)_{j=1...m, i=1...k}$, are forwarded to the encoder \textit{(B)}.
			The first part of this encoder is a feed-forward network \textit{(B1)}.
			The outputs $\bm{u}$ of this network are forwarded to a scaler \textit{(B2)}, that returns a prediction for mean parameters $\hat{\p}$.
			These parameters serve as an input to the decoder \textit{(C)}, which also takes the locations of the electrodes and the points in time, at which the sEMG shall be predicted, $\bm{\mathrm{T}} = \left(\x^j, t^i\right)_{j=1...n_\mathrm{E}, i=1...k}$ as inputs.
			The decoder then predicts the sEMG $\left(\emg(\x^j, \hat{\p}, t^i)\right)_{j=1...n_\mathrm{E}, i=1...k}$ resulting from the parameters $\hat{\p}$ \textit{(C1)}.
			From this predicted sEMG, double differences $ \left(\Delta\emg(\x^j, \hat{\p}, t^i)\right)_{j=1...m, i=1...k}$ are calculated \textit{(C2)} and scaled \textit{(C3)}.
			These scaled sEMG predictions $\bm{\mathrm{N}} =  \left(\Delta {\emg}^*(\x^j, \hat{\p}, t^i)\right)_{j=1...m, i=1...k}$ are then forwarded to the calculation of the loss \textit{(D)}.
			Here, the deviation from the preprocessed sEMG recordings $\bm{\mathrm{M}}$ is computed, forming the loss for training the feed-forward network in the encoder.
		}
		\label{fig:encoderXarchitecture}
	\end{figure*}
	
	However, neural networks, in general, lack interpretability \cite{zhang2021survey}.
	Specifically in this scenario, features learned in the latent space of an autoencoder cannot be ensured to adhere to physically meaningful parameters.
	To ensure their interpretability, an informed variant that overcomes this limitation by integrating domain knowledge directly into the training process is used.
	This variant is inspired by informed autoencoder architectures such as the Encoder-X architecture \cite{wang2021encoder}, which targets the estimation of coefficients for fitting implicit and explicit polynomials, or the informed autoencoder presented by \citet{Terbuch2023} for solving boundary value problems.
	In contrast to classical autoencoder architectures, \citet{Terbuch2023} created a decoder that is not based on a neural network but on a set of admissible functions.
	This set of functions guarantees satisfaction of given boundary constraints and is assumed to be complete.
	The latent variables then form the coefficients of these functions.
	Furthermore, as only the encoder is built by a neural network, the number of weights that need to be trained is smaller than in a classical autoencoder, leading to reduced computational efforts.
	
	This idea is transferred to the non-linear inverse problem that arises when estimating motor unit parameters from sEMG recordings (see Figure \ref{fig:encoderXarchitecture}).
	The encoder part of this architecture is a neural network that calculates latent variables.
	These latent variables serve as motor unit parameters that are given as an input to the decoder.
	This decoder is an explicit sEMG forward model \cite{mechtenberg2023method}, that -- given motor unit parameters $\hat{\p}$ such as the location of the innervation zone centre or the conduction velocity of the electric potential -- predicts the EMG signal resulting from a single muscle fibre at skin level.
	The discrepancy between this prediction and the measured signal is used for training the weights of the encoder.
	After training, the encoder part of the model is used to estimate the motor unit parameters.
	
	The following sections describe the preprocessing of the data (Section \ref{sec:preprocessing}), the architecture of the encoder (Section \ref{sec:encoder}), and the steps of the decoder (Section \ref{sec:decoder}).

	\subsection{\textit{(A)} Preprocessing of sEMG recordings} \label{sec:preprocessing}

	As an array of mono polar electrodes is assumed, double differences $\Delta {\Psi}$ between neighbouring electrodes are calculated by
	\begin{equation} \label{eq:doubleDifferences}
		\Delta {\Psi}(\x^{j}, {\p}, t) = \Psi(\x^{j+2}, {\p}, t) - 2 \cdot \Psi(\x^{j+1}, {\p}, t) + \Psi(\x^{j}, {\p}, t)
	\end{equation}
	for all $j \in \{1, ..., n_\mathrm{E}-2\}$ \cite{mechtenberg2023method}.
	Then, the data are filtered using an 8-th order Butterworth bandpass filter with critical frequencies of $4 \,\si{\hertz}$ and $400 \,\si{\hertz}$.
	Subsequently, each channel of sEMG recordings is scaled to values between $0$ and $1$ using a Min-Max-Scaler.
	The matrix of scaled, preprocessed sEMG recordings is denoted by $\bm{\mathrm{{M}}} = \left(\Delta {\Psi}^*(\x^j, \p, t^i)\right)_{j=1...m, i=1...k}$ with $m = n_\mathrm{E} -2$.
	
	\subsection{\textit{(B)} Encoder} \label{sec:encoder}

	\subsubsection{\textit{(B1)} Feed-forward network}
	
	As with the complete architecture, the design of the encoder part is inspired by Encoder-X \cite{wang2021encoder}.
	A feed-forward network $f(\bm{\mathrm{M}}, \theta)$ with network weights $\theta$, consisting of several blocks of activation layers and normalisation layers (AN blocks), encodes the preprocessed sEMG recordings $\bm{\mathrm{M}}$.
	For the choice of the activation function in the AN blocks (ReLU or leaky ReLU) and the number $n_l \in \mathbb{N}$ of AN blocks (1 to 3), hyperparameter optimization is employed.
	The number of neurons in the first AN block is given by $2^{n_l+2}$ and is halved in each consecutive AN block.
	The number of neurons in the last layer is given by the number of parameters that are to be estimated.
	For this last layer, Encoder-X uses a linear layer, as the parameter space for the polynomial fitting problem is infinite \cite{wang2021encoder}.
	However, as feasible intervals are known for the motor unit parameters, the last layer of the encoder presented in this article is given by a sigmoid activation layer, compressing the outputs of the encoder ${\bm{u}}$ to the range between 0 and 1, allowing for a subsequent scaling to an interval of physically plausible values.

	\subsubsection{\textit{(B2) Scaler}}  
	The outputs ${\bm{u}}$ are scaled such that the parameters take on physically plausible values, e.g. such that the conduction velocity of the electric potential takes on a value between $3$ and $6 \,\si{\metre\per\second}$.
	In the following, this scaling is denoted by function $\bm{g}\left(\bm{u}\right) = \hat{\p}$, where $\hat{\p}$ denotes the scaled parameters that are predicted by the encoder.
	The exact values for the physical scaler can be found in Section \ref{sec:appendix_physical_scaler} in the appendix.
	
	\subsection{\textit{(C)} Decoder} \label{sec:decoder}
	
	The original autoencoder architecture uses a neural network for decoding \cite{hinton2006reducing}; the Encoder-X architecture \cite{wang2021encoder} adapts this framework and uses explicit or implicit polynomials, respectively.
	Then, the reconstruction error is calculated, allowing for the adaptation of the weights of the encoder network.
	In contrast to the Encoder-X architecture, the decoder in this work is formed by a domain model that determines the myoelectric potential  of the electrical field resulting from the activation of a muscle fibre \cite{mechtenberg2023method}.
	This forward model needs to allow for the simulation of sEMG recordings in a short time and be end-to-end differentiable, such that it can be integrated with the training process of the encoder.
	The model published by \citet{mechtenberg2023method} relies on a purely analytical solution, which can be calculated in a short time.
	Furthermore, it is publicly available {\cite{mechtenberg2023githubsimulation}} and already implemented in TensorFlow \cite{tensorflow2015-whitepaper}, allowing for an easy integration with neural networks.
	
	\subsubsection{\textit{(C1)} Forward Model}
	
	The forward model approximates the EMG at skin level resulting from a muscle fibre $i$ by a function ${h} : \mathbb{R}^3 \times \mathbb{R}^q \times \mathbb{R}_{\geq 0} \times \mathbb{R} \rightarrow \mathbb{R}$.
	This function assigns an electrical voltage to a location $\x \in \mathbb{R}^3$ at time $t \in \mathbb{R}_{\geq 0}$, given parameters $\p^i \in \mathbb{R}^q$ specifying the muscle fibre.
	However, as every point along a single muscle fibre emits an electric potential field, the electrical potential $\emg^i$ at a position $\x \in \mathbb{R}^3$ is given by the integral along the muscle fibre.
	Formally,
	\begin{equation}\label{eq:actionPotential1mf}
		\emg^{i} (\bm{x}, \p^i, t) = \int_{0}^{l^i} {h}(\bm{x}, \p^{i}, t, z) dz,
	\end{equation}
	where $z$ integrates over the length $l^i \in \mathbb{R}_{>0}$ of the muscle fibre.
	As a motor unit consists of up to hundreds of muscle fibres, the electric potential of a motor unit denoted as $\Psi (\bm{x}, {\p}, t)$, is the result of the superposition of the potentials [Eq. \eqref{eq:actionPotential1mf}] of $n \in \mathbb{N}$ individual muscle fibres.
	Hence,
	\begin{equation*}
		\Psi (\bm{x}, \p, t) = \sum_{i = 1}^{n} \emg^i(\bm{x}, \p^i, t)
	\end{equation*}
	where $\p = (\p^1, \p^2, ..., \p^n)$ is the aggregated parameter vector.
	However, as this would require the estimation of $n$ parameter vectors $\bm{p}^i$, a simplification is proposed:
	Instead of using $n$ parameter vectors $\p^i$ for $n$ muscle fibres, a \textit{mean muscle fibre} characterised by mean parameters $\hat{\p}$ which approximate the EMG from all muscle fibres of a motor unit is sought, so 
	\begin{equation*} \label{eq:actionPotentialnmf}
		{\emg} (\bm{x}, \hat{\p}, t) =  \int_{0}^{\overline{l}} {h}(\bm{x}, \hat{\p}, t, z) dz.
	\end{equation*}
	The hypothesis underlying this assumption is that the shape of the EMG ${\emg}$ resulting from the mean muscle fibre resembles the shape of the EMG resulting from all muscle fibres of a motor unit, even though their absolute magnitude differs largely.
	
	In accordance with the sEMG recordings of a single motor unit $\left(\Psi(\x^1, \p, t^1), ..., \Psi(\x^{n_E}, \p, t^k)\right)$, the function ${\emg}$ is evaluated at electrode locations $(\x^1, ..., \x^{n_\mathrm{E}})$ at times $t^1, ..., t^k$.
	Consequently, the forward model predicts sEMG recordings $\left(\emg(\x^j, \hat{\p}, t^i)\right)_{j=1...n_\mathrm{E}, i=1...k}$.

	\subsubsection{\textit{(C2)} Double Differentiation}

	Analogously to Eq. \eqref{eq:doubleDifferences}, double differences are calculated from the predicted sEMGs, in the following denoted by $\left(\Delta \emg(\x^j, \hat{\p}, t^i)\right)_{j=1...m, i=1...k}$.
	
	\subsubsection{\textit{(C3)} Scaling}
	
	As the sEMG predicted by the decoder results from a single muscle fibre, its magnitude is smaller than the magnitude of the sEMG recordings resulting from all muscle fibres of the motor unit.
	However, as the target is to compare the shapes of the sEMG recordings instead of their magnitude, a min-max-Scaler scaling each channel to values between $0$ and $1$ is used.
	In the following, $\bm{\mathrm{N}} =  \left(\Delta {\emg}^*(\x^j, \hat{\p}, t^i)\right)_{j=1...m, i=1...k}$ denotes the scaled sEMG double differences resulting from predicted parameters $\hat{\p}$.

	\subsection{\textit{(D)} Loss}

	\begin{figure*}
		\subfloat[\label{fig:plot_lossa}]
		{	
			\includegraphics[width=0.5\linewidth]{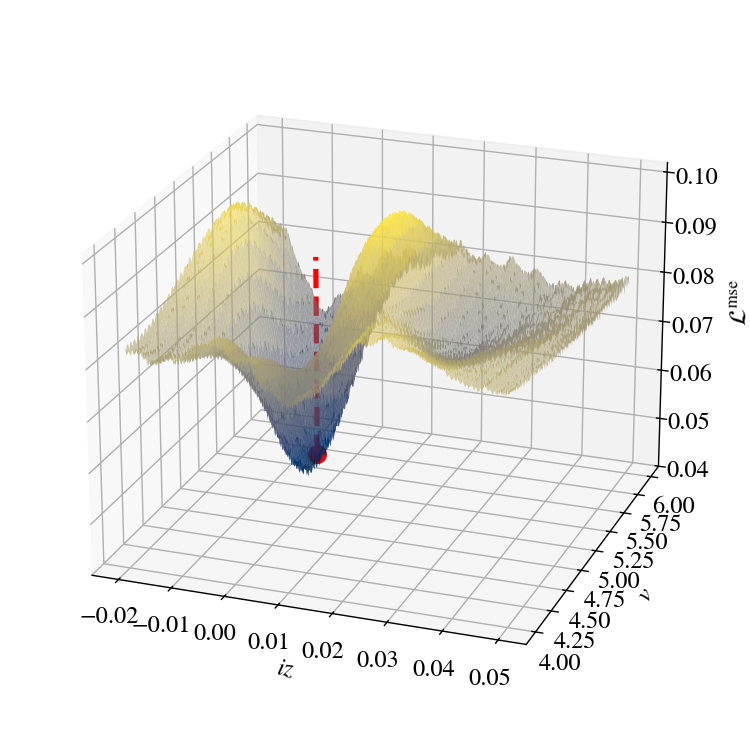}
		}
		\subfloat[\label{fig:plot_lossb}]
		{
			\includegraphics[width=0.5\linewidth]{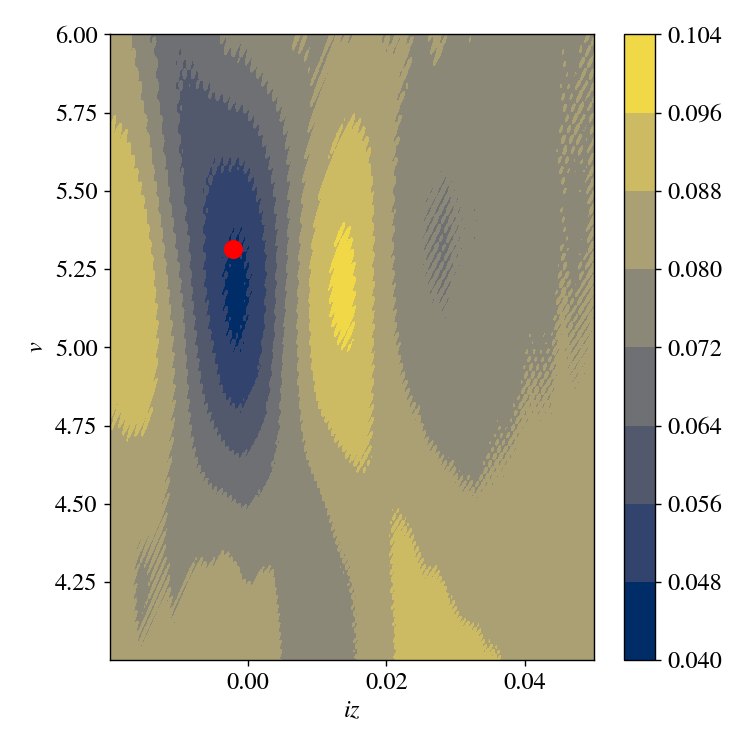}
		}
		\caption{Plot of the MSE loss function $\mathcal{L}^\mathrm{mse}$ for innervation zone centre ${iz}$ and conduction velocity ${v}$ as a 3d surface plot (a) and as a contour plot (b).
			The mean parameters of the motor unit serve as reference values and are marked in red.
		}
		\label{fig:plot_loss}
	\end{figure*}
	
	A parameter vector $\hat{\p}$ minimising the reconstruction error between the double differences of the predicted EMGs ${\bm{\mathrm{N}}}$ and the double differences of the actual measurements  ${\bm{\mathrm{M}}}$ is sought.
	A common metric for measuring the discrepancy between measurements is the mean squared error (MSE).
	Another metric that has been used in the context of estimating motor unit parameters is cross-correlation \cite{Beck2012-fj}.
	In order to measure the discrepancy between EMG recordings and predictions, a loss function combining the MSE and the cross correlation between the two EMGs is employed.
	The mean squared error is calculated by
	\begin{equation}\label{eq:defMSE}
		\mathcal{L}^\mathrm{mse}(\hat{\p})
		= \frac{1}{k \cdot m} \sum_{i=1}^{k} \sum_{j=1}^{m} \left(\Delta {\emg}^*(\x^j, \hat{\p}, t^i) - \Delta {\Psi}^*(\x^j, \p, t^i) \right)^2.
	\end{equation}
	Figure \ref{fig:plot_loss} shows a plot of the MSE loss function for the innervation zone centre $iz$,  i.e.  the mean value of innervation point locations, and the conduction velocity $v$, i.e. the velocity with which the electric potential traverses trough the tissue.

	As the sEMG recordings considered are discrete and no time lag between the two time series is assumed, the negated mean cross correlation is given by
	\begin{equation}\label{eq:defLossCrossCorr}
		\mathcal{L}^\mathrm{cc}(\hat{\p})
		= - \frac{1}{k \cdot m} \sum_{i=1}^{k} \sum_{j=1}^{m}   \Delta {\emg}^*(\x^j, \hat{\p}, t^i) \cdot \Delta {\Psi}^*(\x^j, \p, t^i).
	\end{equation}
	For the estimation of parameters $\hat{\p}$, a combination of these two metrics is used, so
	\begin{equation} \label{eq:lossCombined}
		\mathcal{L}^\mathrm{comb}(\hat{\p}) = \lambda_1 \mathcal{L}^\mathrm{mse}(\hat{\p}) + \lambda_2 \mathcal{L}^\mathrm{cc}(\hat{\p})
	\end{equation}
	with $\lambda_1, \lambda_2 \in \mathbb{R}_{\geq 0}$.
	This loss is used for the training of the encoder, so for searching network parameters $\theta$.
	After training, the encoder is used for the prediction of the motor unit parameters.

	\section{Experimental Evaluation} \label{sec:experiments}
	
	In order to demonstrate the feasibility of the presented approach, motor unit parameters for several synthetic motor units of synthetic muscles are estimated.
	For this purpose, the simulation from \citet{mechtenberg2023method} is used to generate sEMG recordings resulting from all muscle fibres of a motor unit.
	
	Each muscle fibre of a motor unit is parametrised by an innervation point, i.e. the location where the activating neuron is connected to the muscle fibre.
	The approach aims to predict the {innervation zone centre} $\hat{iz}$ and the \textit{conduction velocity} $\hat{v}$, so $\hat{\p} = \left( \hat{iz}, \hat{v} \right)^T$.
	The EMG of this mean muscle fibre should, in its shape, reconstruct the measured EMG; specifically, parameters of this mean muscle fibre are expected to be close to the mean values of all muscle fibres of a motor unit.

	\subsection{Data Generation}
	
	For the experimental evaluation, the EMG signal of six synthetic muscles is simulated.
	Each muscle contains 774 motor units, from which eight specific motor units of different sizes are extracted.
	These motor units contain between 315 and 3367 muscle fibres.
	The conduction velocity is drawn from a normal distribution with mean values between $2.5$ and $5.4 \,\si{\meter\per\second}$ and a standard deviation of 0.22.
	The length of each muscle fibre follows a symmetric triangular distribution over the ranges $[0.145, 0.155]\,\si{\meter}$ and $[0.185, 0.195]\,\si{\meter}$, respectively.
	The cross-sectional area of the muscles is $0.001 \,\si{\square\meter}$.
	
	The sEMG is simulated for a linear electrode array with 40 electrodes, spread out over a distance of $0.195 \,\si{\meter}$, resulting in an inter electrode distance of $5 \,\si{\milli\meter}$.
	$195$ time steps with a frequency of $5000 \,\si{\hertz}$ are recorded, leading to a total time of $39 \,\si{\milli\second}$.
	
	In order to obtain realistic measurements, Gaussian noise is added to the recordings.
	However, to achieve an interpretable signal-to-noise ratio, Gaussian noise is added after double differentiation of the measurements and before scaling.
	A signal-to-noise ratio of $1.0 \,\si{\decibel}$ is chosen.

	\subsection{Implementation Details}
	
	For the training, $10000$ epochs are performed using AdamW Optimizer \cite{loshchilov2017decoupled} with clipnorm $1.0$.
	Furthermore, the training stops early when the loss has not improved by at least $10^{-6}$ over $3000$ epochs.
	For the final estimation, the model with the minimal loss is used.
	The weights of the combined loss [Eq. \eqref{eq:lossCombined}] are set to $\lambda_1 = 0.875$ and $\lambda_2 = 0.125$.
	
	\subsection{Evaluation Metrics}
	
	As a baseline for the parameters, a \textit{prototype muscle fibre} (prot. MF) is created.
	Its parameters (denoted by $iz^\mathrm{PR}$ and $v^\mathrm{PR}$) are given by the mean values of the parameters of all muscle fibres of the regarded motor unit, so
	\begin{equation*}
		iz^\mathrm{PR} = \frac{1}{n} \sum_{i=1}^{n} iz^i
	\end{equation*}
	for the location of the innervation zone and
	\begin{equation*}
		v^\mathrm{PR} = \frac{1}{n} \sum_{i=1}^{n} v^i
	\end{equation*}
	for the conduction velocity.
	
	The sEMGs resulting from the activation of the prot. MF (denoted by $\emg^\mathrm{PR}$) are compared to the actual measurements $\Psi$.
	The MSE and the cross correlation (Eqs. [\eqref{eq:defMSE}, \eqref{eq:defLossCrossCorr}]), calculated on $\emg^\mathrm{PR}$ and $\Psi$, serve as a measure for the systematic error that is made due to the simplification of using a single muscle fibre.
	
	Based on the prot. MF, the following metrics are calculated:
	The absolute error for the innervation zone centre is calculated by $\Delta^\mathrm{abs}_{iz} = |iz^\mathrm{PR} - \hat{iz}|$, i.e. the distance between innervation zone centre of the prot. MF and the estimated value $\hat{iz}$;
	the absolute error for the conduction velocity is calculated by $\Delta^\mathrm{abs}_{v} = |v^\mathrm{PR} - \hat{v}|$, i.e. the absolute difference of the conduction velocity of the prot. MF and the estimated value $\hat{v}$.

	\subsection{Comparison to State-of-the-Art-Method} \label{sec:baseline}
	
	As a baseline for the estimation of the innervation zone centre, the approach from \citet{Mechtenberg2024} is used.
	This approach is based on a combination of tracking motor unit potentials, tracing them back in time, and finding intersections of adjacent motor unit potentials.
	The hypothesis underlying this algorithm is that these intersections are close to the actual innervation points.
	Hence, clustering on the inferred innervation points is performed to locate the innervation zone centre.
	
	The parametrisation of the approach is listed in Section \ref{sec:appendix_clustering}.
	
	\subsection{Results}
	
	\begin{table*}
		\centering	
		\caption{
			Comparison of clustering-based and informed encoder estimation approaches for eight different motor units of one exemplary muscle.
			The estimation of the innervation zone with the lowest absolute error is highlighted in bold.
			The clustering-based algorithm is used for the estimation of the innervation zone centre; hence, only the absolute error of the estimation of the innervation zone is compared.
			Numerical values have been rounded to four decimal places.
			The last four columns contain the MSE loss and the cross correlation loss, once for the predicted parameters $\hat{\p}$, and once for the prot. MF using mean parameters $\p^\mathrm{PR}$.
		}
		\label{tab:results_one_muscle}
		\begin{tabular}{R{0.055\linewidth}R{0.08\linewidth}R{0.08\linewidth}R{0.1\linewidth}R{0.07\linewidth}R{0.065\linewidth}R{0.07\linewidth}R{0.085\linewidth}R{0.07\linewidth}R{0.085\linewidth}R{0.07\linewidth}} 
			\hline 
			\multirow[c]{2}{\linewidth}{Motor Unit ID }& \multirow[c]{2}{\linewidth}{${iz}^\mathrm{PR}$ in \,\si{\milli\metre} }& \multirow[c]{2}{\linewidth}{${v}^\mathrm{PR}$ in \,\si{\metre\per\second}}& \multirow[c]{2}{\linewidth}{Estimation Method }& \multirow[c]{2}{\linewidth}{${\Delta}^\mathrm{abs}_{iz}$ in \,\si{\milli\metre}}& \multirow[c]{2}{\linewidth}{${\Delta}^\mathrm{abs}_{v}$ in \,\si{\metre\per\second}}& \multicolumn{2}{c}{prediction $\hat{\p}$}&\multicolumn{2}{c}{prot. MF ${\p}^\mathrm{PR}$} 	\\ &&&&&& {${\sqrt{\mathcal{L}^\mathrm{mse}}}$ in \,\si{\volt} }& {${{\mathcal{L}^\mathrm{cc}}}$ in \,\si{\square\volt} }& {${\sqrt{\mathcal{L}^\mathrm{mse}}}$  in \,\si{\volt} }& {${{\mathcal{L}^\mathrm{cc}}}$  in \,\si{\square\volt}} \\  \cline{7-10}
			\hline
			\multirow{2}{*}{0} & \multirow{2}{*}{8.4808} & \multirow{2}{*}{3.9852} & Clustering & \textbf{1.4362} & -- & -- & -- & -- & --  \\
			\cline{4-10}
			&  &  & Informed AE  & 1.6487 & 0.0397 & 0.2271 & 0.1738 & 0.2304 & 0.1745 \\
			\hline
			\multirow{2}{*}{1} & \multirow{2}{*}{-4.4375} & \multirow{2}{*}{4.1820} & Clustering & \textbf{0.8603} & -- & -- & -- & -- & --  \\
			\cline{4-10}
			&  &  & Informed AE  & 2.9215 & 0.0327 & 0.2088 & 0.1719 & 0.2152 & 0.1710 \\
			\hline
			\multirow{2}{*}{2} & \multirow{2}{*}{-10.0267} & \multirow{2}{*}{4.3922} & Clustering & {0.5533} & -- & -- & -- & -- & --  \\
			\cline{4-10}
			&  &  & Informed AE  & \textbf{0.0238} & 0.2658 & 0.2434 & 0.1831 & 0.2504 & 0.1821 \\
			\hline
			\multirow{2}{*}{3} & \multirow{2}{*}{-7.3191} & \multirow{2}{*}{4.5750} & Clustering & \textbf{1.7906} & -- & -- & -- & -- & --  \\
			\cline{4-10}
			&  &  & Informed AE  & 5.4130 & 0.0897 & 0.2250 & 0.1868 & 0.2317 & 0.1863 \\
			\hline
			\multirow{2}{*}{4} & \multirow{2}{*}{-7.5293} & \multirow{2}{*}{4.7402} & Clustering & \textbf{0.9479} & -- & -- & -- & -- & --  \\
			\cline{4-10}
			&  &  & Informed AE  & 3.7964 & 0.0172 & 0.2384 & 0.1938 & 0.2436 & 0.1903 \\
			\hline
			\multirow{2}{*}{5} & \multirow{2}{*}{0.2393} & \multirow{2}{*}{4.9449} & Clustering & 1.3856 & -- & -- & -- & -- & --  \\
			\cline{4-10}
			&  &  & Informed AE  & \textbf{0.2983} & 0.0447 & 0.2245 & 0.2076 & 0.2266 & 0.2055 \\
			\hline
			\multirow{2}{*}{6} & \multirow{2}{*}{0.6045} & \multirow{2}{*}{5.1253} & Clustering & \textbf{0.4972} & -- & -- & -- & -- & --  \\
			\cline{4-10}
			&  &  & Informed AE  & 2.3884 & 0.0036 & 0.1985 & 0.1958 & 0.2019 & 0.1970 \\
			\hline
			\multirow{2}{*}{7} & \multirow{2}{*}{-2.2305} & \multirow{2}{*}{5.3135} & Clustering & \textbf{0.5197} & -- & -- & -- & -- & --  \\
			\cline{4-10}
			&  &  & Informed AE  & 0.8015 & 0.0812 & 0.2218 & 0.1988 & 0.2243 & 0.1994 \\
			\hline
		\end{tabular}
	\end{table*}
	
	For all motor units examined, the training converges.
	After training, the encoder is used to predict the innervation zone centre and the conduction velocity.
	These values are compared to the mean parameters of the prot. MF; the results are shown in Tables \ref{tab:results_one_muscle} and \ref{tab:results_agg_mu}.

	Table \ref{tab:results_one_muscle} shows the comparison of the clustering-based algorithm and the informed AE for one exemplary muscle.
	For two out of eight motor units, the informed autoencoder (AE) achieves a lower absolute error than the clustering-based approach (Section \ref{sec:baseline}) when estimating the innervation zone centre; for the remaining six motor units, the clustering-based approach achieves a lower absolute error.
	In total, the informed AE is capable of estimating the innervation zone centres with absolute errors of at most $\sim\,5.5 \,\si{\milli\meter}$, in six out of eight cases with absolute errors below $3 \,\si{\milli\meter}$.
	The conduction velocities are estimated with absolute errors below $0.3 \,\si{\meter\per\second}$, in seven out of eight cases with absolute errors below $0.1 \,\si{\meter\per\second}$.
	
	\begin{table*}
		\centering	
		\caption{
			Comparison of clustering-based and informed encoder estimation approaches for several muscles.
			The results are aggregated over the estimations for each of the eight motor units of the corresponding muscle.
			Numerical values have been rounded to four decimal places.
		}
		\label{tab:results_agg_mu}
		\begin{tabular}{
				R{0.06\linewidth}
				R{0.08\linewidth}
				R{0.08\linewidth}
				R{0.1\linewidth}
				R{0.06\linewidth}
				R{0.07\linewidth}
				R{0.07\linewidth}
				R{0.085\linewidth}
				R{0.07\linewidth}
				R{0.085\linewidth}
				R{0.07\linewidth}}
			\hline \multirow{2}{\linewidth}{Muscle ID }& \multirow{2}{\linewidth}{$\overline{iz}^\mathrm{PR}$ in \,\si{\milli\metre} }& \multirow{2}{\linewidth}{$\overline{v}^\mathrm{PR}$ in \,\si{\metre\per\second}}& \multirow{2}{\linewidth}{Estimation Method }& \multirow{2}{\linewidth}{$\overline{\Delta}^\mathrm{abs}_{iz}$ in \,\si{\milli\metre}}& \multirow{2}{\linewidth}{$\overline{\Delta}^\mathrm{abs}_{v}$ in \,\si{\metre\per\second}}& \multicolumn{2}{c}{prediction $\hat{\p}$}&\multicolumn{2}{c}{prot. MF ${\p}^\mathrm{PR}$} 	\\&&&&&&$\overline{\sqrt{\mathcal{L}^\mathrm{mse}}}$ in \,\si{\volt} & $\overline{{\mathcal{L}^\mathrm{cc}}}$ in \,\si{\square\volt} & $\overline{\sqrt{\mathcal{L}^\mathrm{mse}}}$  in \,\si{\volt} & $\overline{{\mathcal{L}^\mathrm{cc}}}$  in \,\si{\square\volt}\\\hline$\multirow{2}{*}{0}$ & $\multirow{2}{*}{-2.7773}$ & $\multirow{2}{*}{4.6573}$ & Clustering & $0.9988$ & $--$ & $--$ & $--$ & $--$ & $--  $\\
			\cline{4-10}
			&  &  & Informed AE  & $2.1614$ & $0.0718$ & $0.2234$ & $0.1889$ & $0.2280$ & $0.1883 $\\
			\hline
			$\multirow{2}{*}{1}$ & $\multirow{2}{*}{-0.7870}$ & $\multirow{2}{*}{4.6573}$ & Clustering & $0.8554$ & $--$ & $--$ & $--$ & $--$ & $--  $\\
			\cline{4-10}
			&  &  & Informed AE  & $3.3441$ & $0.3480$ & $0.2487$ & $0.1952$ & $0.2470$ & $0.1972 $\\
			\hline
			$\multirow{2}{*}{2}$ & $\multirow{2}{*}{-1.7775}$ & $\multirow{2}{*}{4.6573}$ & Clustering & $1.1784$ & $--$ & $--$ & $--$ & $--$ & $--  $\\
			\cline{4-10}
			&  &  & Informed AE  & $3.1199$ & $0.1609$ & $0.2507$ & $0.2103$ & $0.2529$ & $0.2105 $\\
			\hline
			$\multirow{2}{*}{3}$ & $\multirow{2}{*}{0.0448}$ & $\multirow{2}{*}{4.5635}$ & Clustering & $0.3949$ & $--$ & $--$ & $--$ & $--$ & $--  $\\
			\cline{4-10}
			&  &  & Informed AE  & $2.6004$ & $0.1824$ & $0.2383$ & $0.1952$ & $0.2387$ & $0.1961 $\\
			\hline
			$\multirow{2}{*}{4}$ & $\multirow{2}{*}{-1.2341}$ & $\multirow{2}{*}{4.6573}$ & Clustering & $0.9362$ & $--$ & $--$ & $--$ & $--$ & $--  $\\
			\cline{4-10}
			&  &  & Informed AE  & $2.6289$ & $0.1379$ & $0.2563$ & $0.2011$ & $0.2626$ & $0.2002 $\\
			\hline
			$\multirow{2}{*}{5}$ & $\multirow{2}{*}{-0.5169}$ & $\multirow{2}{*}{4.6573}$ & Clustering & $1.2268$ & $--$ & $--$ & $--$ & $--$ & $--  $\\
			\cline{4-10}
			&  &  & Informed AE  & $1.7387$ & $0.1170$ & $0.2462$ & $0.2101$ & $0.2493$ & $0.2109 $\\
			\hline
			\hline\multirow{2}{*}{mean} & \multirow{2}{*}{-1.1747} & \multirow{2}{*}{4.6417} & Clustering & 0.9317 & -- & -- & -- & -- & --  \\
			\cline{4-10}
			&  &  & Informed AE  & 2.5989 & 0.1697 & 0.2439 & 0.2001 & 0.2464 & 0.2005 \\
			\hline
		\end{tabular}
	\end{table*}
	
	Table \ref{tab:results_agg_mu} shows the aggregated results for six different muscles.
	For each of these six muscles, parameters for eight motor units were estimated.
	The values in the table are mean values aggregated over the eight motor units, denoted by $\overline{(\,\cdot\,)}$.
	In all cases, when estimating the innervation zone centre the mean absolute error is below $3.5 \,\si{\milli\meter}$.
	The mean absolute error when estimating the conduction velocity is always below $0.35 \,\si{\meter\per\second}$, in seven out of eight cases with absolute errors below $0.2  \,\si{\meter\per\second}$.
	
	\subsection{Interpretation of Results}
	
	In most cases, the estimations of the clustering-based algorithm is closer to the innervation zone centre than the estimation of the informed AE.
	However, it can be seen that in some scenarios the estimations of the informed AE are closer than those of the clustering-based algorithm.
	Furthermore, the informed AE for some motor units reaches even smaller losses than that of the prot. MF, from which the estimation errors are calculated (e.g. motor units 1 and 2).
	This implies that the parameters of the prot. MF are not always in the global minimum of the loss function (independently of the weighting parameters $\lambda_1$ and $\lambda_2$ of the loss function [Eq. \eqref{eq:lossCombined}]).
	As the training minimizes a combination of the MSE and the cross correlation, a more exhaustive training would not necessarily lead to a solution closer to the prot. MF.
	Consequently, future work will examine an adaptation of the loss function to further reduce the estimation errors.

	\section{Conclusion and Outlook} \label{sec:conclusion}
	
	This work presents a novel approach for the estimation of motor unit parameters from sEMG recordings.
	An informed autoencoder architecture is presented that allows for the simultaneous estimation of several parameters whilst respecting physical constraints.
	For this purpose, a feed-forward network serving as the encoder is integrated with a forward model predicting sEMG recordings from motor unit parameters, which serves as the decoder.
	The encoder is then trained to predict those parameters that minimise the deviation between the predicted and the actual sEMG recordings.
	The presented model reduces manual modelling efforts and can be adapted to predict several motor unit parameters simultaneously:
	given that a parameter is included in the forward model and that a suitable set of data to estimate the parameter is available, the presented approach is capable of estimating this parameter.
	
	Experiments on eight different motor units of six muscles show, that the approach is capable of estimating the innervation zone centre with a mean absolute error of $2.5989 \,\si{\milli\meter}$ and of estimating the conduction velocity of the electric potential with a mean absolute error of $0.1697 \,\si{\meter\per\second}$.
	
	Future work will focus on refining the approach to further increase the accuracy of parameter estimation.
	Furthermore, the generalisation capabilities of the approach, i.e. whether a neural network can be pre-trained on several sEMG recordings and subsequently fine-tuned for a specific individual, will be evaluated.

	\appendix
	
	\section{Technical Appendix}\label{sec:appendix}

	\subsection{Parameters of Physical Scaler}\label{sec:appendix_physical_scaler}
	
	The estimation of the innervation zone centre is scaled with the distance between the first and the last electrode.
	The hypothesis underlying this scaling is that the innervation zone centre lies somewhere between the electrodes.
	The estimation of the conduction velocity is scaled such that it lies between $3$ and $6 \,\si{\meter\per\second}$.
	
	\subsection{Parameters for Clustering-based Estimation of Innervation Zone Centre} \label{sec:appendix_clustering}
	
	Following \citet{Mechtenberg2024}, the wavelet width is set to $\lambda = -0.00391667$; the density parameter of clustering algorithm DBSCAN  is set to $\varepsilon = 1.10083333$.
	The reference conduction velocity used for scaling is set to $5 \,\si{\meter\per\second}$.
	
	\subsubsection*{Declaration of Generative AI Use}
	
	During the preparation of this work the authors used generative AI for improving formulations as well as spelling and grammar checks.
		
\end{document}